\documentclass[sigplan,nonacm]{acmart}

\usepackage[]{hyperref}
\usepackage{xcolor}

\usepackage{subcaption}

\newcommand{\sysname}{Pie}

\begin{document}

\title{\sysname: Pooling CPU Memory for LLM Inference}

\author{Yi Xu, Ziming Mao, Xiangxi Mo, Shu Liu, Ion Stoica}

\affiliation{%
  \institution{UC Berkeley}
  \city{}
  \country{}
}

\begin{abstract}
The rapid growth of LLMs has revolutionized natural language processing and AI analysis, but their increasing size and memory demands present significant challenges. A common solution is to spill over to CPU memory; however, traditional GPU-CPU memory swapping often results in higher latency and lower throughput.

This paper introduces \sysname{}, an LLM inference framework that addresses these challenges with performance-transparent swapping and adaptive expansion. By leveraging predictable memory access patterns and the high bandwidth of modern hardware like the NVIDIA GH200 Grace Hopper Superchip, \sysname{} enables concurrent data swapping without affecting foreground computation, expanding effective memory without added latency. Adaptive expansion dynamically adjusts CPU memory allocation based on real-time information, optimizing memory usage and performance under varying conditions.

\sysname{} maintains low computation latency, high throughput, and high elasticity. Our experimental evaluation demonstrates that \sysname{} achieves optimal swapping policy during cache warmup and effectively balances increased memory capacity with negligible impact on computation. 
With its extended capacity, \sysname{} outperforms vLLM by up to 1.9$\times$ in throughput and 2$\times$ in latency. Additionally, \sysname{} can reduce GPU memory usage by up to 1.67$\times$ while maintaining the same performance. Compared to FlexGen, an offline profiling-based swapping solution, \sysname{} achieves magnitudes lower latency and 9.4$\times$ higher throughput.
\end{abstract}

\maketitle
\pagestyle{plain}

\section{Introduction}

In recent years, the application of Large Language Models (LLMs) has seen widespread adoption, becoming a cornerstone of modern technology. These models are essential for a range of applications, from natural language processing to sophisticated AI-driven analysis. However, efficient memory management for LLM inference remains a challenge.

LLMs are growing in size~\cite{yang2023harnessing, kaplan2020scaling}, and so are the prompts they process. These models typically have billions of parameters and work with massive datasets and key-value caches when generating responses. The maximum allowed context length of commercially available LLMs has been increasing exponentially~\cite{anthropic2024claude, achiam2023gpt, google2024gemini}. While efforts have been made to increase GPU memory capacity, the compute capacity continues to grow faster than the memory capacity.

When memory demands exceed GPU memory, a natural solution is to spill over to CPU memory, solution also known as swapping. Unfortunately, swapping can lead to higher latency and lower throughput, as the GPU may need to wait for data spilled over to the CPU memory to be transferred back.
While increasing the CPU memory capacity allows us to process larger data batches, the added wait time may results in both higher latency and lower throughput.

With LLM inference, per-token latency consists of two parts: computation latency and queuing latency. Computation latency is the time it takes to compute the token,  
while queuing latency is the time the token has to wait because it cannot be scheduled as soon as it arrives, typically due to the system being fully utilized.
A larger GPU memory capacity can achieve lower end-to-end latency because of lower queuing latency, and higher throughput because of increased processing capacity. Therefore, the optimal swapping solution would be to add more memory capacity without affecting per-token computation latency (i.e., without causing the GPU to wait for data transfers). Achieving this will result in lower end-to-end per-token latency and higher throughput.

In this paper, we ask the following questions: "What is the maximum CPU memory we can add without blocking GPU computation?" and "What is the performance impact of extending capacity using CPU memory?" We explore these questions by designing \sysname{}, an LLM inference framework that based on \textit{performance-transparent swapping} and \textit{adaptive expansion}. Performance-transparent swapping ensures that swapping has zero impact on the compute latency. 
Adaptive expansion dynamically allocates the optimal amount of CPU memory for swapping, maintaining performance transparency and preventing resource under-utilization.

At its core performance-transparent swapping allows for the swapping process to occur concurrently with foreground computation. 
Performance-transparent swapping gives the application the illusion of a GPU with more memory without any negative impact on latency. Note this is different from virtual memory; while virtual memory also gives the illusion of more memory, it comes at the cost of latency due to paging when the data is not in memory. Performance-transparent swapping improves performance by using a larger memory without the associated latency penalties. 

Therefore, by adjusting the amount of swapping, the system can modify the effective memory size. \textit{Effective memory} is defined as the maximum amount of memory that the application can use with no impact on the GPU performance. 
It includes the total available GPU memory plus \textit{effective extended memory}, which is a certain amount of dynamically allocated CPU memory that can be used without blocking GPU computation. This approach effectively increases the memory size while introducing a degree of elasticity.

To achieve true performance transparency in swapping, data required for upcoming accesses must be prefetched to the compute device before those accesses occur. The prefetch bandwidth imposes a strict constraint on the data size that can be transferred between devices within a given time interval. 
The prefetch efficiency measures the fraction of prefetched data that turns out to be useful and can be utilized without blocking the GPU.
The effective extended memory size is bounded by the product of the prefetch bandwidth, the duration of the prefetch time interval, and the prefetch efficiency.
Therefore, the swapping mechanism relies on high-bandwidth devices and must accurately predict upcoming memory accesses.

There are two properties that  
facilitate performance transparent swapping. First, recent GPUs, such as NVIDIA's GH200 Grace Hopper Superchip~\cite{nvidia2024gracehopper}, provide high-bandwidth interconnect between the GPU and CPU memories. By leveraging NVLink, this bandwidth can be as high as 900GB/s. Second, the memory accesses of the LLM inference workloads are predictable, especially at the layer granularity.
This predictability allows us to achieve 100\% prefetching efficiency.
Together, these two properties allow us to substantially increase the effective memory for LLM inference, potentially extending this memory by tens of gigabytes.

Determining the appropriate amount of CPU memory for swapping is a non-trivial task. If the allocated size is too large, \sysname{} will break the transparency of swapping, causing applications to experience delays as computations are blocked waiting for data. Conversely, if the size is too small, resources may be underutilized, leading to lower performance. Additionally, this decision is influenced by workload characteristics; some workloads do not benefit from increased memory because they are compute-bound and require only a small amount of memory. 
Adding more memory to such workloads would be a waste of resources.

One technique to optimize the performance of LLM inference is using offline profiling: we profile and decide the memory allocations \emph{before} running the application.
Unfortunately, offline profiling ignores changes in the workloads  and the system environment during  the run time (e.g., another process might initiate data transfer during the inference workload). 

To address this challenge, \sysname{} employs adaptive expansion, a lightweight online method, to determine the amount of CPU memory for swapping. It starts with zero CPU memory  and then gradually increases this memory as long as the following conditions: (1) the GPU-CPU interconnect is not saturated; (2) the swapping latency remains lower than computation latency; (3) the workload throughput increases as we allocate more CPU memory. 
If any of these conditions are not met, \sysname{} maintains the current allocation or reduces the amount of CPU memory allocated for swapping.

This paper makes four key contributions:
\begin{itemize}
    \item We introduce performance-transparent swapping, enabling LLM inference systems to use CPU memory without blocking computation.
    \item We present adaptive expansion, a technique that dynamically adjusts the swapping size to accommodate changes in the system environment and workload, achieving optimal performance under varying conditions.
    \item We design and implement a system prototype, \sysname{}, that achieves high throughput, low latency, and high elasticity.
    \item We provide a thorough experimental evaluation showing that \sysname{} outperforms vLLM by up to $1.9\times$ in throughput and $2\times$ in latency. Compared to FlexGen, \sysname{} achieves up to $60\times$ lower latency and $9.4\times$ higher throughput.
\end{itemize}

The rest of this paper is organized as follows: Section~\ref{sec:background} gives background on LLM inferences and motivates \sysname{}. In Section~\ref{sec:design}, we discuss the design, and in Section~\ref{sec:impl}, we detail the implementation. Section~\ref{sec:eval} presents \sysname{}'s performance. We discuss related work in Section~\ref{sec:related} and conclude in Section~\ref{sec:conclusion}.

\section{Background}
\label{sec:background}
This section discusses the context of LLM inference, highlighting memory as one of the most important bottlenecks. It reviews existing solutions to this issue and introduces the GH200 architecture as a promising new hardware to overcome this challenge, discussing the trade-offs involved in using the device.

\subsection{LLM inference}
LLMs have demonstrated significant capabilities in tasks such as chatting, content creation, and programming. Deploying these applications requires GPU-based systems to run the LLMs. However, serving LLMs remains slow and costly, requiring many high-end GPUs for production-level services.

Today's LLMs primarily operate on autoregressive Transformer models \cite{vaswani2017attention}, generating words as tokens sequentially based on a given prompt and previously produced tokens. A token represents a condensed version of a character series, typically encoded using Byte-Paired Encoding \cite{sennrich2015neural}, with an average token equating to about four English characters.

The inference mechanism for LLMs is split into two phases: (i) the prefill stage, where the model processes all input prompt tokens simultaneously, and (ii) the decoding stage, where it sequentially generates tokens. This token generation is sequential because it depends on the accumulated conditional probabilities of all preceding tokens, continuing until a termination token is produced.
An LLM inference engine, such as vLLM \cite{kwon2023efficient}, TGI \cite{huggingface_tgi}, or TensorRT-LLM \cite{nvidia_trt_llm}, executes transformer models and orchestrates the prefill and decode stages. Since each request must be processed sequentially, the engine batches multiple requests together \cite{yu2022orca} to boost throughput. Efficient memory management is crucial, as the intermediate state for all tokens in a batch must be stored in the key-value (KV) cache simultaneously.

\subsection{Memory bottleneck}

\paragraph{KV Cache Benefits from Larger Size.}
The KV cache size increases significantly with more requests. For OPT-13B, with a per-token KV cache of 800 KB, one request with 2048 tokens requires 1.6 GB. Modern GPUs, with memory in the tens of GBs, can process only a limited number of requests simultaneously. Increasing the memory size allows LLM serving systems to expand the batch size, thereby improving both throughput and latency.

\paragraph{KV Cache is Dynamic.}
LLM memory management must dynamically adapt to changing workloads and system environments.
The varying input and output sizes of LLM workloads require a flexible memory system to handle diverse prompt lengths. 
Rapid fluctuations in workload characteristics require real-time adjustments~\cite{zheng2023judging, taori2023stanfordalpaca, sharegpt}, 
while changes in system environments, such as variations in bandwidth or computational resources, also impact memory usage efficiency. 
The necessity to balance memory utilization and maintain performance stability under varying conditions makes LLM memory management more challenging.

\subsection{Existing Solutions}

Recent advancements in LLM inference have led to the development of dedicated systems tailored to this domain. Examples include FasterTransformer~\cite{nvidia_fastertransformer}, Orca~\cite{yu2022orca}, vLLM~\cite{kwon2023efficient}, FlexGen~\cite{sheng2023flexgen} and so on. Orca is a LLM inferece system that demonstrates significant throughput improvements for LLM inference by iteration-level scheduling techniques. 
vLLM is a system built based on PagedAttention mechanism, allows for even more efficient memory management and higher throughput inference by storing attention keys and values in non-contiguous paged memory.

FlexGen addresses the constraints of limited GPU memory by offloading the computational and memory demands of LLM inference to a combination of GPU, CPU, and disk resources. By optimizing the storage and access patterns of tensors and employing weight and cache compression, FlexGen extends the capabilities of conventional hardware setups and provides solutions for systems with limited memory.

These systems use swapping mechanisms that expose lower performance devices to applications. Many employ a best-effort strategy to transfer necessary data to the GPU before it is needed. For instance, in vLLM, requests may be preempted and swapped out when memory capacity is insufficient.
Once memory becomes sufficient again, these requests need to be rescheduled and swapped back in. During this time, vLLM must wait for the swap-in process to complete before it can proceed with serving the requests. 
Similarly, FlexGen stores model weights, attentions, and activations on the CPU. It triggers a swap-in only when certain data is needed and not found in the GPU. These on-demand swap-ins often cause delays, as computations must wait for data transfers to complete.
As a result, applications in these systems experience increased latency and reduced throughput while waiting for data transfers from CPU memory or disk to GPU memory. 

Furthermore, these systems lack the capability to dynamically adapt to changing workloads or system environments, potentially leading to resource under-utilization, higher latency, and lower throughput.

\subsection{Graph Hopper Superchip}

The NVIDIA GH200 Grace Hopper Superchip~\cite{nvidia2024gracehopper}, integrating Hopper GPU architecture and Grace CPU, is tailored for demanding AI and high-performance computing workloads. 
This advanced superchip is equipped with 96GB of HBM3e memory, capable of delivering memory performance with a substantial bandwidth of 4 terabytes per second. 
Another GH200 standout features is NVIDIA's NVLink Chip-2-Chip interconnect, which facilitates high-speed, high-bandwidth communication and coherence between the CPU and GPU. 
It achieves a data transfer rate of 419 GB/s from CPU to GPU and 371 GB/s from GPU to CPU. 

\begin{figure}
    \centering
    \includegraphics[width=0.8\linewidth]{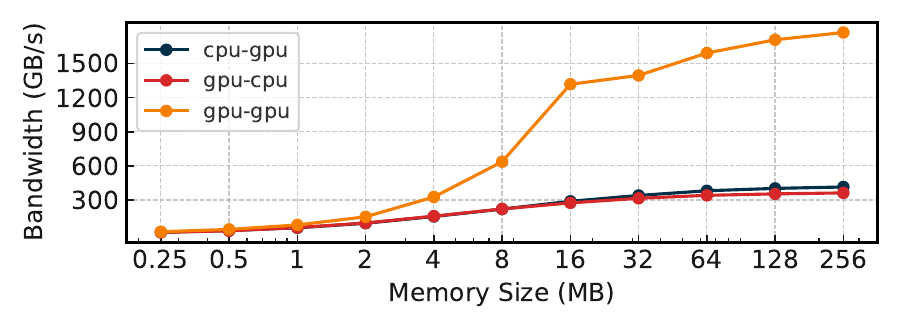}
    \caption{GH200 Bandwidth Measurement}
    \label{fig:gh200}
\end{figure}

We measured the CPU-GPU bandwidth using nvbandwidth~\cite{nvbandwidth}, with the results shown in Figure~\ref{fig:gh200}. We found that:
1. Access granularity is important, with only accesses larger than 16MB utilizing 95\% of the memory bandwidth. However, in real applications, GPUs can hide slowdowns from small accesses through their memory hierarchy and memory coalescing. In contrast, directly accessing the CPU exposes much lower bandwidth.
2. The effective bandwidth between CPU and GPU is only about 1/4 $\times$ of the GPU-GPU bandwidth.
These demonstrate the need of accessing CPU memory through the GPU using a swapping-like mechanism.

\section{Design}
\label{sec:design}

\sysname{} employs performance-transparent swapping, operating layer by layer through the KV cache. By prefetching data for upcoming layers through a FIFO queue, \sysname{} masks the latency of memory swapping, overlapping it with ongoing computations, thus expanding the effective GPU memory capacity. 
\sysname{} also uses adaptive expansion to optimize capacity expansion, dynamically adjusting expansion size based on real-time workloads and system conditions to ensure efficient resource utilization while maintaining performance transparency.

We provide an abstracted logical view of using the CPU memory pool as effective memory extension in \ref{subsec:vir-gpu}, present \sysname's architecture in \ref{subsec:pi-architecture}, describe the design of performance-transparent swapping in \ref{subsec:transparent-swapping}, and discuss how adaptive expansion achieves optimal swapping policies in \ref{subsec:adaptive-expansion}.

\subsection{Expand Effective GPU Memory Size}
\label{subsec:vir-gpu}

In this subsection, we describe our methods for GPU memory expansion, specifically focusing on the memory allocated for KV cache during LLM inference.

Our model considers both a physical and a logical view of GPU memory. Physically, the GPU designates $a$ GB for KV cache, with the remaining capacity used for model weights and a variety of intermediate results. 
Assuming the model has $n$ layers, each with a corresponding KV cache, results in $n$ layers of KV cache. We relocate $m$ layers of the KV cache to the CPU, leaving $n-m$ layers on the GPU.
\sysname{} uses per-layer KV cache as the basic management unit, making it simple, model-agnostic, and non-intrusive.
 
The capacity for each layer is defined as the total KV cache capacity divided by the number of logical layers.
Initially, $a$ GB of GPU memory was divided among $n$ layers, resulting in a capacity of \(\frac{a}{n}\) GB per layer. When $m$ layers are transferred to the CPU, the GPU's memory originally allocated for these $m$ layers can now be redistributed among the remaining $(n-m)$ layers, enhancing their capacity to \(\frac{a}{n-m}\) GB each. Assuming the CPU's capacity is underutilized, the expanded GPU KV cache effectively expands to a total capacity of $n \times \frac{a}{n-m}$ GB. We describe this extension in capacity using the term \textit{expansion}, defined as the ratio $\frac{n}{n-m}$. In subsequent sections, expansion denotes the increased capacity ratio.

The value of expansion is crucial to the performance.
Ideally, expansion should be maximized because a higher expansion leads to a larger effective memory size. The batch size, or the number of requests that can be concurrently served, is determined by the available memory size. Therefore, this increase in effective memory size increases the batch size, potentially increasing throughput and reducing end-to-end latency.

But expansion comes with a cost. Because higher expansion means more layers are going to be physically on CPU, which, in return, means more data need to be transferred during inference. 
In order to realize performance-transparent swapping, which requires swapping latency to be hidden by the already existed computation latency, $t_{\text{swap}} \times m$ must be equal or less than $t_{\text{compute}} \times n$. 
Here, $t_{\text{compute}}$ represents the time taken to compute one layer, and $t_{\text{swap}}$ represents the time taken to swap one layer between the CPU and GPU.

Per-layer computation latency does not increase linearly with the KV cache size, but per-layer swapping latency does. 
Higher expansion also results in more layers residing on the CPU at any given time. Consequently, more layers are involved in swapping during each token's generation.

Initially, without swapping, the swapping latency among all layers is zero. As expansion increases, both per-layer swapping and computation increase, more layers are involved in swapping. This combined effect causes the total swapping latency to increase much faster than the total computation latency.
The maximum tolerable number of layers involved in swapping is reached when the computation and swapping latencies become equal, or $t_{swap} \times m = t_{compute} \times n$.

\subsection{\sysname{} Architecture}
\label{subsec:pi-architecture}

\begin{figure*}
    \centering
    \includegraphics[width=0.8\linewidth]{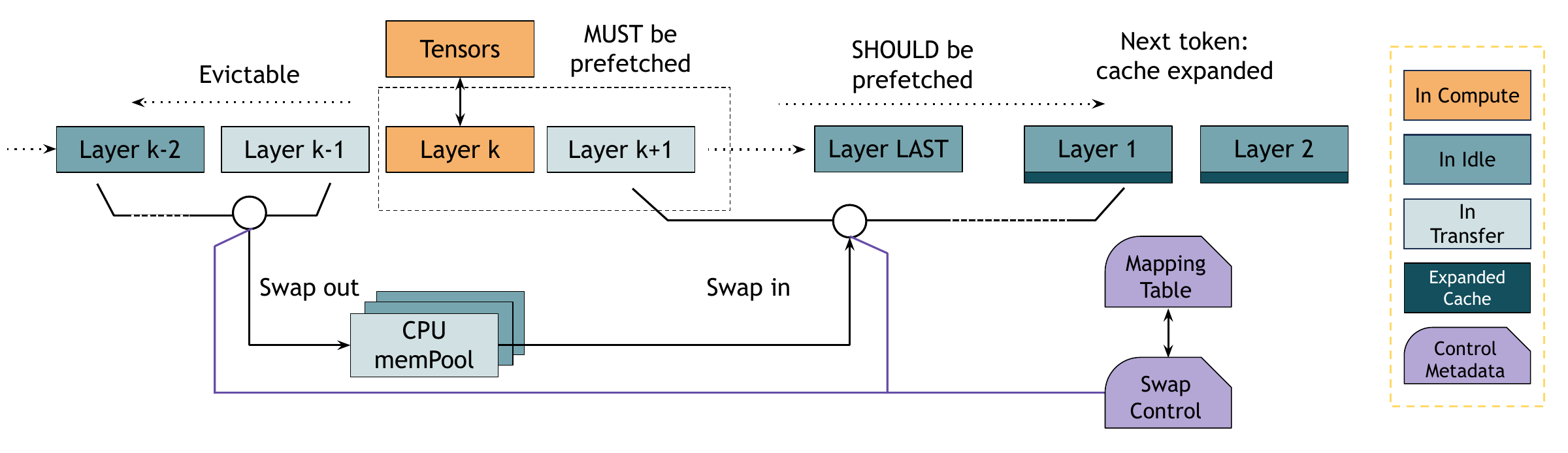}
    \caption{\sysname{} Architecture.}
    \label{fig:pi-arch}
\end{figure*}

In the inference process of LLMs, data is processed token by token and then layer by layer. At any given time, the GPU tensors interact only with the KV cache of the current layer. After spending a computation time \( t_{\text{compute}} \) processing the current layer \( k \), the tensors begin processing the next layer \( k+1 \). This processing uses the output from layer \( k \) and the KV cache updated from the previous token iterations for layer \( k+1 \). The updated cache for each layer is accessed at intervals of \(\text{n} \times t_{\text{compute}}\).

Performance-transparent swapping happens concurrently as the computation progress through the layers.
Ideally, swapping out the cache of layers that are not in use to CPU memory and reloading it before the phase in which they are needed in the next iteration should not impede the inference process.

In the \sysname{} architecture, the cache states of recently completed layers and upcoming layers are clear. The recently completed layers (Layer \(k-2\), Layer \(k-1\) in Figure~\ref{fig:pi-arch}) are swappable, meaning they can be swapped out to CPU memory because they won't be in use for a while. The caches for upcoming layers (Layer \(k\), Layer \(k+1\) in Figure~\ref{fig:pi-arch}) must be prefetched to GPU memory before they are accessed. For layers a bit further ahead (Layer \(\text{LAST}\) in Figure~\ref{fig:pi-arch}), they should be prefetched if CPU-GPU interconnect is available.
However, the caches states of other layers remains uncertain: if they were swapped out and are not immediately needed, they free up GPU memory space and can be swapped back in later. If they have remained in GPU memory due to limited swapping bandwidth, they do not occupy transfer bandwidth and can be directly used in the next iteration.

To maximize effective cache size through performance-transparent swapping, an optimal policy should minimize the number of swaps, as swapping is expensive, consuming bandwidth and causing a logical layer to occupy memory on both the CPU and GPU simultaneously. This can be achieved by maximizing the time each swapped-out layer spends in CPU memory, as longer durations lead to fewer swaps and lower overhead.

\sysname{} features a swapping control logic, as illustrated in Figure~\ref{fig:pi-arch}. 
Based on the current computation layer index and a mapping table of per-layer KV caches, it dynamically determines whether a layer should be swapped out or if a previously swapped-out layer needs to be swapped back in at a given time. To prevent delays in computation, it is crucial that the caches of the swapped-out layers are fully transferred back via the CPU-GPU interconnect before the computation begins.

In the \sysname{} architecture, the caches of each layer do not have a fixed allocation; the KV cache is not statically partitioned between the CPU and GPU. 
In other words, the logical layers do not have a fixed physical per-layer cache on either CPU or GPU; their physical per-layer cache is dynamically allocated during swap-in/out operations.

\begin{figure}
    \centering
    \includegraphics[width=0.9\linewidth]{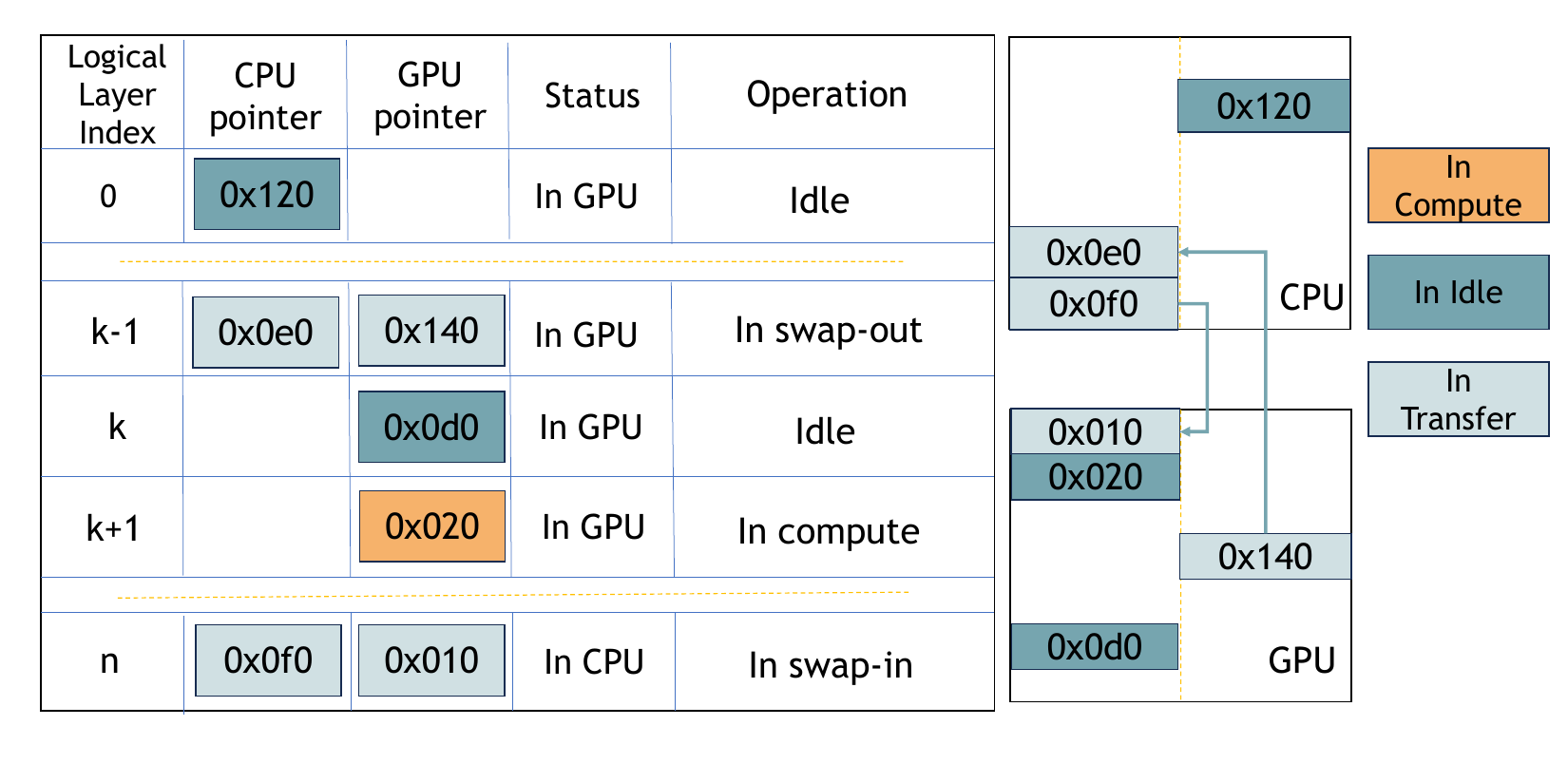}
    \caption{Mapping Table used in Swapping Control, current time=003.}
    \label{fig:mapping-table}
\end{figure}

\sysname{} employs a mapping table within its swapping control logic, functioning similarly to a traditional page table in an operating system. This mapping table maintains the correspondence between the logical layer index and its location in either the CPU or GPU cache, as illustrated in Figure~\ref{fig:mapping-table}.

As swapping begins, the swapping control identifies eligible layers for swapping and transfers them to CPU memory, updating the mapping table. CPU pointers facilitate cache retrieval back to the GPU, while GPU pointers manage actual cache data access. Timestamps are also recorded in the table.

During the swapping process, the mapping table must be updated in a two-phase manner. First, memory is allocated to the layer, and the CPU pointer is stored in the table, making it accessible to the swapping logic. The swapping then begins and proceeds in the background. Only when the swap operation is confirmed to be completed is the mapping table updated to reflect that the layer is physically present on the destination device.

Before computation at layer $k$ begins, \sysname{} must check if the layer is physically present on the GPU by dereferencing the GPU pointer. If the pointer does not exist or the swapping operation has not finished, the computation must wait until the layer has been completely moved back to the GPU.

\subsection{Performance-transparent Swapping}
\label{subsec:transparent-swapping}

Swapping policy is crucial for effective swapping performance. To enable performance-transparent swapping, it is essential that the time each token spends on swapping does not exceed its compute time. An optimal swapping policy maximizes the expansion value, effectively extending memory capacity. Conversely, a suboptimal policy results in significantly lower expansion values, which wastes memory bandwidth and limits potential memory enhancements, or delays computing tasks, thereby violating the goal of performance transparency.

In an ideal setup, precise profiling of each model, hardware, and workload combination is feasible. We can then estimate the swapping latency for different data transfer sizes and the computation latency for each token. Based on these estimations, our goal is to identify the maximum data transfer size, denoted as \(b\) GB, that ensures the swapping latency remains within the computation latency limits. Assuming \(a\) GB of memory is available on the GPU for KV cache usage, this approach effectively expands the memory size to \(a+b\) GB.
The expansion factor, previously defined as \( \frac{n}{n-m} \) and used to compare the logical number of layers to those physically on the GPU, now extends to model the effective memory size relative to the physical GPU memory size, previously denoted as \( \frac{a+b}{a} \). 
Each device must reserve additional space for swap in/out buffers, resulting in 4 buffers in the entire effective memory space and 2 buffers on the physical GPU. The ratio is refined to \( \frac{a+b}{a} = \frac{n+4}{n-m+2} \). 
Given the known values of \(a\), \(b\), and \(n\), we can then compute the value of \(m\).

In this scenario, the optimal protocol is straightforward. When a layer begins computation, it first checks if the previous swap-in has completed. If the swap-in has finished, the system initiates a swap-in for the "hottest layer", which refers to the layer currently in the CPU that is closest to being accessed next. Similarly, the system checks if the previous swap-out has completed. If the swap-out has finished, the system initiates a swap-out for the "coldest layer", defined as the last layer that finished computation. This choice ensures that all other layers' values will be accessed before this layer is needed again. An exception to the swap-out protocol occurs if every layer currently in the CPU is colder than the coldest layer on the GPU. In such cases, the swap-out operation does not proceed.

\begin{figure}
    \centering
    \includegraphics[width=0.99\linewidth]{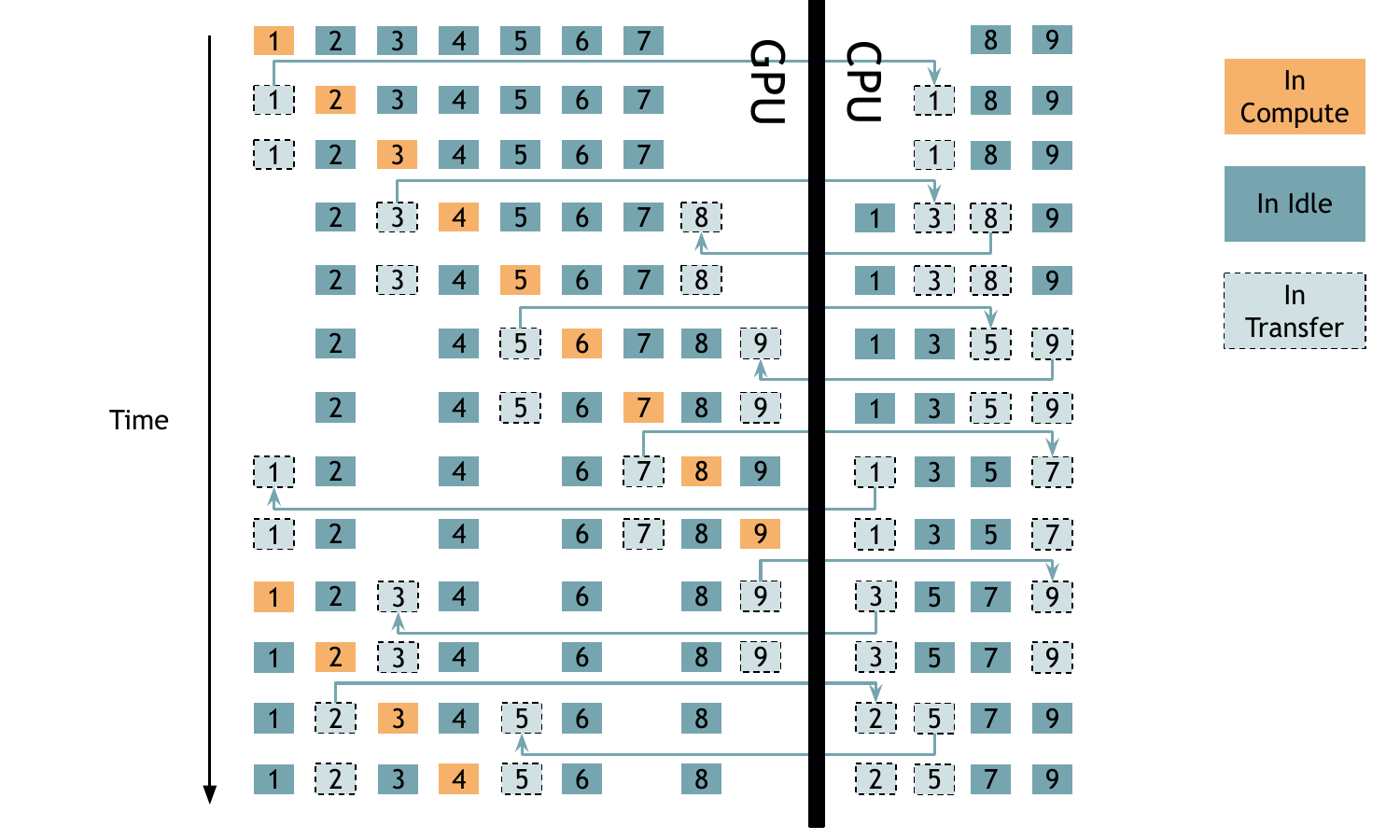}
    \caption{Optimal Swapping Policy}
    \label{fig:optimal-swapping-policy}
\end{figure}

Figure~\ref{fig:optimal-swapping-policy} shows a nine-layer model where swapping time for each layer is twice the computation time. The model allocates two layers in the CPU, with an expansion factor of 1.28. After the initial warm-up, both CPU and GPU use the memory space of two layers as a swapping buffer, acting as sender and receiver for swap-in and swap-out operations.

In this example, a swap-out does not occur during the computation of layer 1 because the CPU layers (8 and 9) are colder than any GPU layers. During layer 2's computation, layer 1 becomes the coldest, and its swap-out is initiated. Since there is no available space for a swap-in, this operation is skipped. During layer 3's computation, layer 2 becomes the coldest, but because layer 1's swap-out is still in progress, both swap-out and swap-in operations are skipped. At layer 4, layer 3 is identified as the coldest, allowing its swap-out. With layer 1 now on the CPU, its original GPU space is allocated to swap in layer 8, the hottest layer on the CPU.

Following this protocol ensures that each layer chosen to be swapped to the CPU is the coldest across both the CPU and GPU at the moment of initiating the swap-out. Since the layer to be swapped out is the logically coldest, the swap-out always occurs after the previous swap-out has finished. After the initial warm-up, there will not be a condition where no layer currently in the CPU is hotter than the coldest layer on the GPU. When swapping out layer $k_n$, it is ensured that previously swapped-out layers $k_p$, $k_q$, and $k_m$ are growing hotter in the order they were swapped out: the earlier a layer was swapped to the CPU, the hotter it becomes.

Therefore, the optimal swapping protocol essentially functions as a First In, First Out (FIFO) queue. The time each layer spends on the CPU is maximized to an optimal value bounded by the swapping latency. 
As illustrated in Figure~\ref{fig:optimal-swapping-policy}, each layer spends the equivalent of eight layer computation times on the CPU. 
The FIFO queue is four layers long because the expansion ensures two layers are physically on the CPU and two are in swapping. Each layer takes the equivalent of two layer computation times to swap in, waits for two preceding layers in the FIFO queue to be processed, and another two layer computation times to swap out. Thus, a layer spends a total of eight layer computation times in the FIFO queue.

\subsection{Adaptive Expansion}
\label{subsec:adaptive-expansion}

\subsubsection{Deciding Optimal Expansion is Challenging}

Determining the optimal expansion value is a complex task. Each combination of model, workload, and hardware has unique computation and swapping latencies, which influence the maximum effective memory capacity expansion size. This size, in turn, determines the optimal expansion ratio. However, accurately identifying this value requires thorough consideration, as computation and swapping latencies can vary even on the same machine with the same model. For instance, larger batch sizes, resulting from either larger cache sizes or shorter contexts, and more complex attention mechanisms can lead to longer computation latencies. Similarly, smaller swapping granularity and larger cache sizes may result in increased swapping latencies. More importantly, even thorough profiling can only ensure optimal expansion on an isolated, stable system where swapping and computation latency do not change. This is often not the case, as CPU-GPU communication can occur unpredictably due to concurrent tasks, thereby significantly impacting swapping latency. Thus, precisely matching computation and swapping latencies to achieve optimal expansion is a challenging task.

If the expansion value is estimated to be higher than the optimal value, the swapping of many layers may not be completed before the computation requires access to them. This delay in swapping can block the start of computation, resulting in increased latency and violating the requirement of performance-transparent swapping.

If the expansion value is estimated to be lower than the optimal value, it will not only limit the potential extent of memory expansion but also result in a significant waste of memory bandwidth when following the same swapping policy.
The optimal expansion value is selected to maximize the effective extended memory size within the given time interval. Any value smaller than the optimal will result in a reduced effective extended memory size. However, with the given time interval and a swapping policy that swaps whenever the interconnect is free, smaller expansion values will cause unnecessary data to be swapped out and in.

We could also design other swapping policies to statically generate swap in and out policies for each layer at the beginning of each token's computation. However, this approach is also difficult.
First, this requires precise estimation of both swapping and computation latencies; otherwise, it may result in delays in computation.
Moreover, even with precise estimation, this remains a complex mathematical problem. Each swap decision affects the subsequent swap decisions. Therefore, to achieve the optimal value, the decision complexity is \(\left((m + 1) \times (n - m + 1)\right)^{n}\), where \(m\) is the number of layers in the CPU and \(n-m\) is the number of layers in the GPU. At each layer, it needs to decide whether to swap in a layer among \(m\) layers and whether to swap out a layer among \(n-m\) layers, resulting in \(\left((m + 1) \times (n - m + 1)\right)\) combinations (including the possibility of no swap). Even if we constrain the swap in and out operations to only the layers closest to the current layer (hottest/coldest layers), the complexity is still at least \(4^{n}\). Searching for the optimal value among these combinations is unrealistic. Simpler algorithms may not achieve the best possible results.
Lastly, layers of adjacent tokens are not entirely independent. This further complicates the problem. For example, if the system detects that the CPU-GPU interconnect is free at a given token but decides not to swap in the first layer of the next token, the subsequent token will need to wait at the first layer until the swap-in is completed. On the other hand, if the next token's first layer is swapped in too early, the layer that has been swapped out to the CPU for this space might not be cold enough (e.g., the swap out could wait until layer 30, but the policy might decide to do it at layer 23). This reduces its time in CPU memory, leading to more frequent swaps and higher overhead.

\subsubsection{Adaptively Changing Expansion}
\label{subsubsec:adaptively-changing-expansion}
Therefore, \sysname{} employs an adaptive swapping policy to achieve optimal swapping, as shown in Figure~\ref{fig:optimal-swapping-policy}, without requiring extensive profiling or calculation. More importantly, this policy allows the system to adjust to dynamic workloads. \sysname{} begins with all layers in the GPU cache and then starts the workload, gradually increasing the number of layers in the CPU cache. Assuming the size of available GPU memory remains constant, increasing the number of layers stored on the CPU results in a larger per-layer cache size.

\begin{figure}
    \centering
    \includegraphics[width=0.9\linewidth]{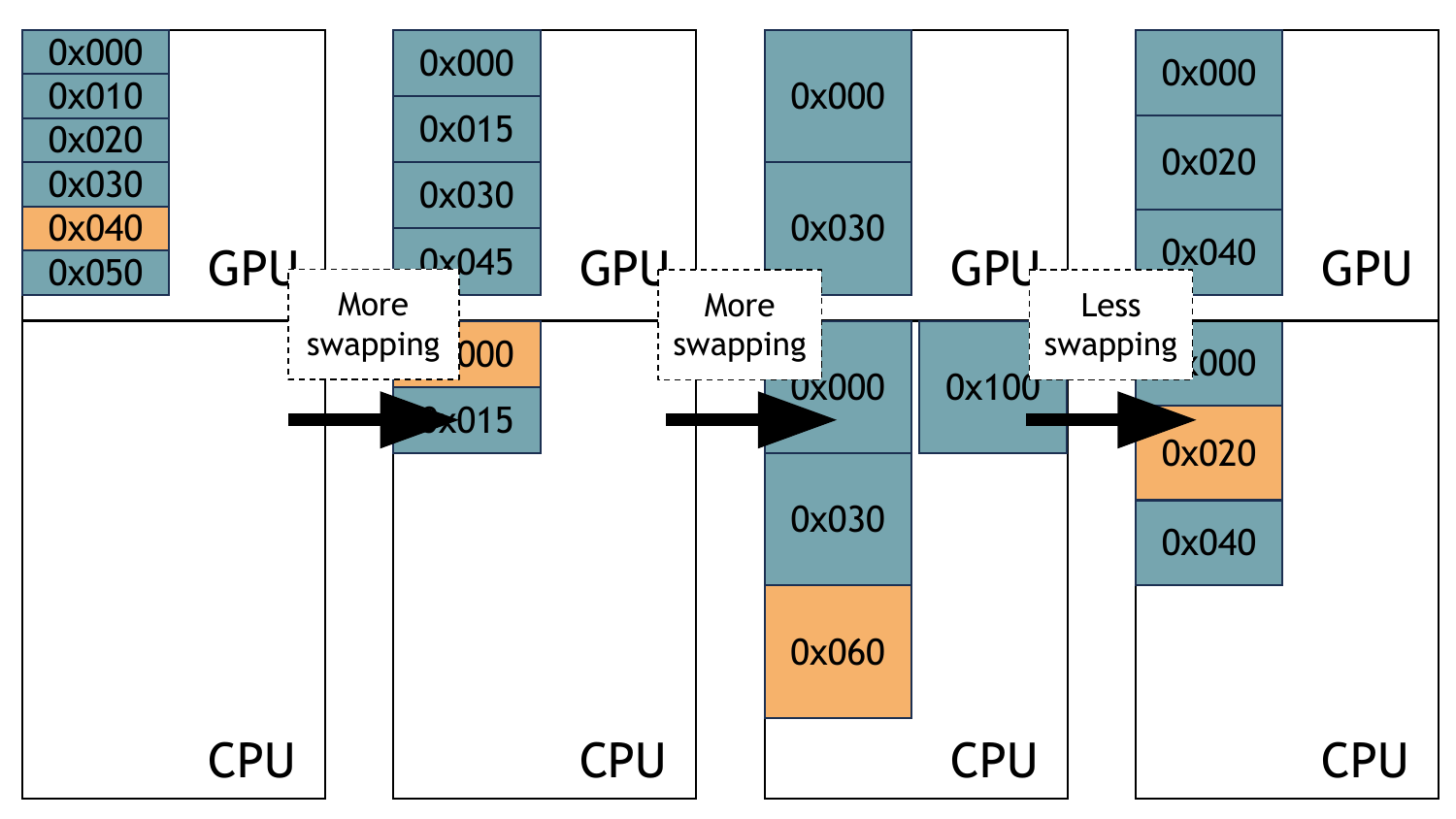}
    \caption{\sysname{} Achieves Optimal Expansion Adaptively. Assuming a 6-layer model, this illustrates the cache status when the 5th layer is in computation, while all other layers are idle (neither in computation nor transfer).}
    \label{fig:swapping-policy-adaptive}
\end{figure}

The condition for \sysname{} to set more layers into swapping is straightforward: at the time of initiating a swap-in or swap-out, \sysname{} checks if additional on-CPU layers are needed for the current token computation. If no on-CPU layers are required, it indicates the current expansion is too small and should be increased, leading to a larger per-layer cache and more layers in the CPU. This triggers the ``More Swapping'' events in Figure~\ref{fig:swapping-policy-adaptive}. 
To prevent waiting, we heuristically allow \sysname{} to swap in the first few layers of the next token. This rule violation occurs at a very low frequency. When the violation hits a preset ratio, it triggers an increase in expansion.

The condition that triggers \sysname{} to reduce swapping with smaller expansion is also straightforward. 
\sysname{} checks whether the swapping latency remains lower than the computation latency. If the model has to wait for swapping to finish before continuing computation, \sysname{} reduces the expansion and moves more layers back to the GPU.
The example in Figure~\ref{fig:swapping-policy-adaptive} shows an example where the latency of swapping four layers per token is too high, causing the computation to wait. Consequently, the number of layers on the CPU is reduced by one.

When adjusting expansion, \sysname{} compares performance records with the current configuration. If the records show that the workload stops scaling with higher expansion, \sysname{} stops increasing expansion and tests if decreasing it has any adverse effects. In scenarios where \sysname{} has more bandwidth but low workload demands, it periodically checks the workload by increasing expansion and reduces it if the workload remains low.

Benefiting from the mapping table, \sysname{} can dynamically adjust the expansion. It quickly converges to the optimal expansion at the beginning of a static workload and reacts promptly to any changes in the background traffic that reduces the effective bandwidth available. An exception might be occasional, unexpected CPU-GPU traffic. In such cases, \sysname{} may experience a temporary slowdown as it waits. If this traffic is prolonged, \sysname{} will adapt accordingly. Otherwise, \sysname{} will tolerate the brief interference without making adjustments.

\section{Implementation}
\label{sec:impl}

\sysname{} is built on vLLM, an end-to-end serving system with a FastAPI~\cite{fastapi_2023} frontend and a GPU-based inference engine. The frontend extends the OpenAI API interface~\cite{openai2020}. We followed the original vLLM implementation for the frontend.

We added 3,000 lines of Python and 100 lines of C++/CUDA code to enable performance-transparent swapping and adaptive expansion. This mostly self-contained code manages the layer mapping table and triggers necessary swaps and expansion changes. Additionally, we developed a CUDA kernel for efficient per-layer KV cache swapping.

\subsection{Support Adaptive expansion}
vLLM manages memory in blocks using a block manager. When a request runs out of memory, the block manager allocates an additional block to it. Once the request is completed or preempted, all allocated blocks are returned to the block manager, which then adds them to a free block list.
To enable adaptive expansion, we modified the block manager to handle dynamically changing per-layer memory spaces.

To increase the per-layer cache size, \sysname{} first prepares the memory space before making them visible. 
After the expansion increment is triggered by the conditions specified in Section~\ref{subsubsec:adaptively-changing-expansion}, \sysname{} initiates a reallocation process. When swapping a layer into a dynamically allocated per-layer cache, if this cache is allocated for the first time to handle swap-in/out operations following the expansion trigger, its size will be incremented by reallocating it to a different location with sufficient contiguous memory.
At this stage, only blocks within the previous size are accessible, and accesses are safely redirected through the mapping table. This redirection is achieved because accesses are performed via a base pointer and offset, with the offset of each block remaining unchanged and the new base pointer obtained through the mapping table.
Once every layer has undergone at least one swap, all layers' spaces have been physically incremented. \sysname{} then adds the newly incremented blocks to the free block lists in the block manager, making them allocatable and accessible.

Similarly, when decreasing the per-layer cache size, \sysname{} reallocates the entire per-layer cache to a new contiguous address during swapping. All blocks remain accessible, but \sysname{} avoids scheduling any requests that use blocks beyond the new decreased size. Once all layers' sizes have been decreased, \sysname{} removes the blocks beyond the new size from the free block lists in the block manager.

\subsection{Support Different Models}
For the model executor, we implemented support for popular LLMs such as OPT~\cite{zhang2022opt} and LLaMA~\cite{touvron2023llama}, which vLLM originally implemented using PyTorch~\cite{paszke2019pytorch} and Transformers~\cite{wolf2020transformers}.
Adding support for different models is straightforward and requires less than 50 lines of code to add a control loop that triggers performance-transparent swapping or adapts to expansion. 
\section{Evaluation}
\label{sec:eval}

In this section, we aim to comprehensively evaluate our system by addressing the following key questions:

\begin{enumerate}
\item How much GPU memory can \sysname{} save? (\S\ref{memory-usage})
\item How much performance improvement can \sysname{} provide compared to vLLM, assuming the same amount of GPU memory is used? (\S\ref{pi-vs-vllm})
\item How does the request rate impact \sysname{}'s performance compared to other systems? (\S\ref{request-rate-scale})
\item In \sysname{}, how important is the choice of expansion? How effective is the self-adaptive swapping in achieving optimal performance? (\S\ref{adaptive-expansion})
\item How does \sysname{}'s design for improving throughput affect the average per token end-to-end latency? (\S\ref{latency-throughput})
\item How does \sysname{} compare to state-of-the-art swapping-based systems? (\S\ref{pi-vs-flexgen})
\end{enumerate}

\subsection{Evaluation Setup}

\paragraph{Model and Server Configurations.}
We evaluated the following models: OPT with 13B and 30B parameters~\cite{zhang2022opt}, and Llama with 13B parameters~\cite{touvron2023llama}. All experiments were conducted on NVIDIA Grace Hopper instances, featuring 480GB of LPDDR5X DRAM on the CPU and 96GB of HBM3 on the GPU, interconnected via 900GB/s NVLink~\cite{nvidia_grace_hopper_spec}.

\paragraph{Workloads.}
We used workloads from ShareGPT~\cite{sharegpt, chatgpt} and Alpaca~\cite{taori2023stanfordalpaca, wang2022self} datasets, containing real LLM service texts. ShareGPT features longer inputs and outputs than Alpaca. 
We generate request arrival times using a Poisson distribution with different request rates. 
We tokenized the datasets following prior work~\cite{kwon2023efficient}.

\paragraph{Metrics}

We measure serving throughput as the total number of output tokens divided by the duration. We also report per-token latency, which includes queuing latency (total wait time divided by output token numbers) and compute latency (total compute time divided by output token numbers).

\paragraph{Baselines}
vLLM~\cite{kwon2023efficient} is an inference engine optimized for managing memory fragmentation, enhancing throughput and latency through paged attention. \sysname{} is built on top of vLLM.
FlexGen~\cite{sheng2023flexgen} is a state-of-the-art swapping-based LLM inference framework. It uses synchronous swapping to fetch data on demand if it is not in GPU memory. FlexGen utilizes offline profiling to fit a cost model for optimal allocation and swapping policies. Based on the hardware setup (CPU/GPU memory, prompt length, output length), the model determines the allocation of weights, attention KV cache, and activations between CPU and GPU.

\subsection{Higher Throughput, Lower GPU Memory Usage}
\label{memory-usage}

\begin{figure*}[ht]
    \centering
    \begin{subfigure}[b]{0.24\textwidth}
        \includegraphics[width=\linewidth]{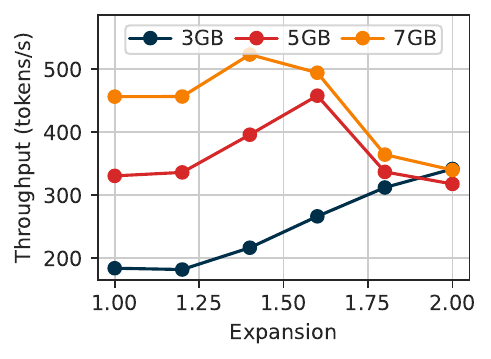}
        \caption{ShareGPT on OPT-30B}
        \label{fig:sub1}
    \end{subfigure}
    \hfill
    \begin{subfigure}[b]{0.24\textwidth}
        \includegraphics[width=\linewidth]{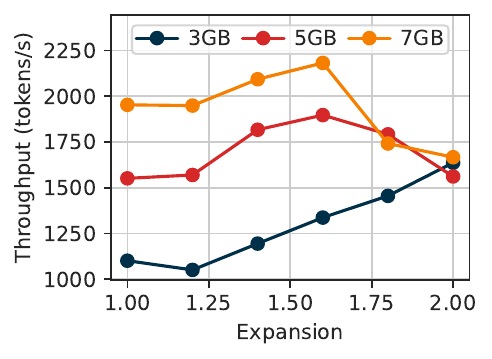}
        \caption{Alpaca on OPT-30B}
        \label{fig:sub2}
    \end{subfigure}
    \hfill
    \begin{subfigure}[b]{0.24\textwidth}
        \includegraphics[width=\linewidth]{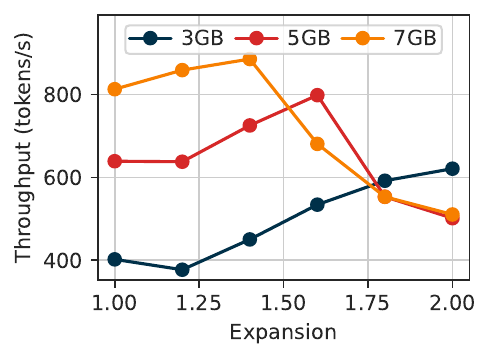}
        \caption{ShareGPT on OPT-13B} 
        \label{fig:sub3}
    \end{subfigure}
     \begin{subfigure}[b]{0.24\textwidth}
        \includegraphics[width=\linewidth]{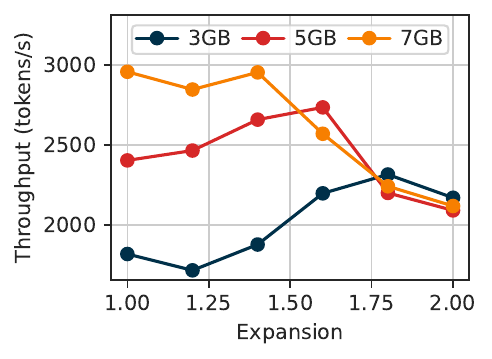}
        \caption{Alpaca on Llama-13B} 
        \label{fig:sub6}
    \end{subfigure}
    \caption{\sysname{} achieves higher throughput with lower GPU memory usage, across various Physical GPU KV Cache sizes. } 
    \label{fig:less-gpu-mem}
\end{figure*}

In LLM inference, throughput and latency do not scale indefinitely with increasing KV cache size. Each combination of workload, model, and hardware has a scaling limit beyond which additional performance gains are not observed, despite unused GPU memory.

Different workloads have varying memory intensity on the same model and hardware. For workloads on GPU, either GPU cores or memory will be fully utilized first. When GPU cores are used up before memory, the workload becomes compute-bound. Many workloads become compute-bound well before exhausting KV cache physical GPU memory. 

Given this constraint, it is more efficient to allocate only the necessary GPU memory for inference tasks, optimizing performance without excessive resource use.
A key question to consider is whether this required memory (n GB) needs to be entirely on the GPU. \sysname{} explores reducing GPU memory usage while maintaining optimal throughput and latency by transparently leveraging CPU memory.

In Figure~\ref{fig:less-gpu-mem}, we run alpaca and shareGPT workloads on OPT-13B, OPT-30B and Llama-13B, across different expansion values.
We observe that increasing expansion improves the model throughput up to a certain point, beyond which the overhead of swapping increase as computation now has to wait for swapping to finish. The turning point where throughput is maximized differs for different models and physical GPU KV cache size combinations. For example, running OPT-13B on the ShareGPT dataset, for 3GB of physical GPU KV cache size, the throughput scales throughout the expansion values from 1 to 2. For 7GB of physical GPU KV cache size, the throughput scales until expansion reaches 1.4 or 1.6, depending on workloads and datasets, beyond which the overhead dominates. 
This highlights the importance of adaptive expansion that dynamically chooses the best expansion value to maximize inference throughput. 

Comparing to the baseline performance where there is no expansion (i.e., expansion $=1$), The optimal expansion value results in a throughput improvement of up to $1.86\times$ when running OPT-30B on the ShareGPT dataset, $1.58\times$ when running OPT-30B on the Alpaca dataset, $1.54\times$ when running OPT-13B on the ShareGPT dataset, and $1.27\times$ when running Llama-13B on the Alpaca dataset respectively.

We also found that with a high expansion value, \sysname{} can match on-device performance while using up to $1.67\times$ less GPU memory. Performance-transparent swapping is able to effectively reduce GPU memory usage by expanding into CPU memory. We observed similar trends across other models and workloads.

\subsection{Performance Gain Over vLLM}
\label{pi-vs-vllm}

\begin{figure}[ht]
    \centering
    \begin{subfigure}[b]{0.23\textwidth}        \includegraphics[width=\linewidth]{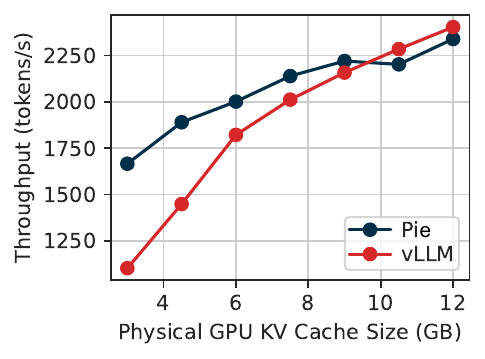}
        \caption{OPT-30B, Alpaca}
        \label{fig:sub1}
    \end{subfigure}
    \hfill
    \begin{subfigure}[b]{0.23\textwidth}
        \includegraphics[width=\linewidth]{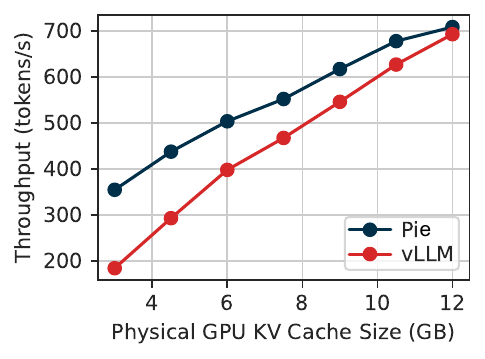}
        \caption{OPT-30B, ShareGPT}
        \label{fig:sub2}
    \end{subfigure}
    \\
    \begin{subfigure}[b]{0.23\textwidth}
        \includegraphics[width=\linewidth]{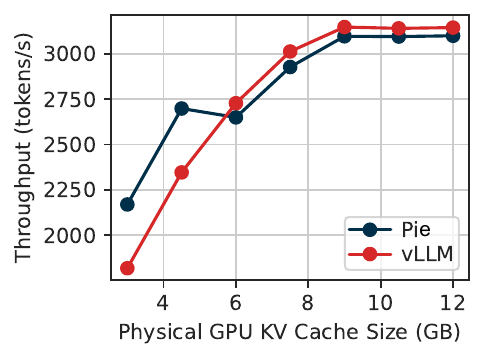}
        \caption{Llama-13b, Alpaca}
        \label{fig:sub2}
    \end{subfigure}
    \hfill
    \begin{subfigure}[b]{0.23\textwidth}
        \includegraphics[width=\linewidth]{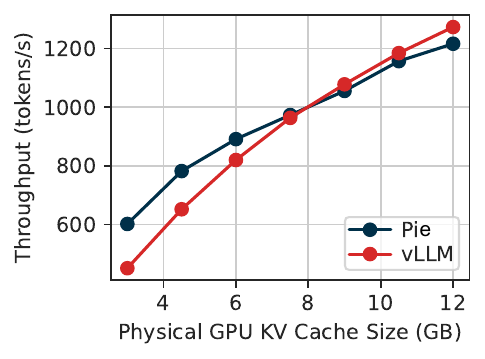}
        \caption{Llama-13b, ShareGPT}
        \label{fig:sub2}
    \end{subfigure}

    \caption{LLM inference throughput with different GPU KV Cache Sizes (GB). } 
    \label{fig:perf-over-vllm}
\end{figure}

\begin{figure}[htbp]
    \centering
    \begin{subfigure}[b]{0.23\textwidth}
        \includegraphics[width=\linewidth]{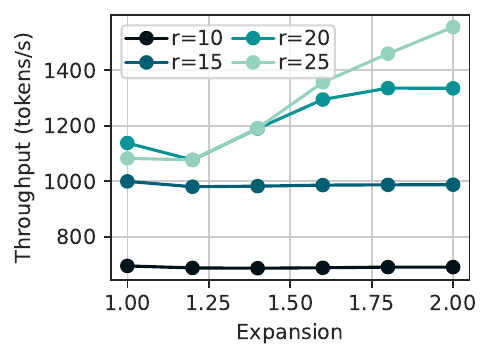}
        \caption{Alpaca}
        \label{opt30-scaling-alpaca}
    \end{subfigure}
    \hfill
    %\\
    \begin{subfigure}[b]{0.23\textwidth}
        \includegraphics[width=\linewidth]{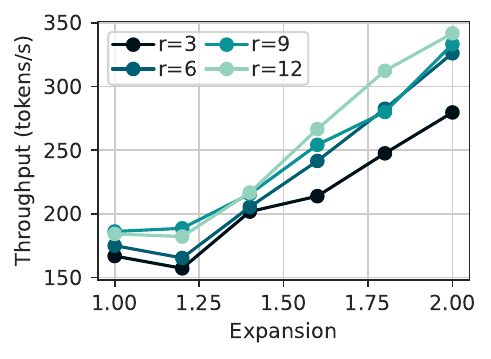}
        \caption{ShareGPT} 
        \label{fig:sub2}
    \end{subfigure}
        \caption{OPT-30B Throughput Change with Expansion on Different Req Rates. Physical GPU KV cache Size = 3.} 
        \label{opt30-scaling}
\end{figure}

\begin{figure*}[ht]
    \centering
    \begin{subfigure}[b]{0.31\textwidth}
        \includegraphics[width=\linewidth]{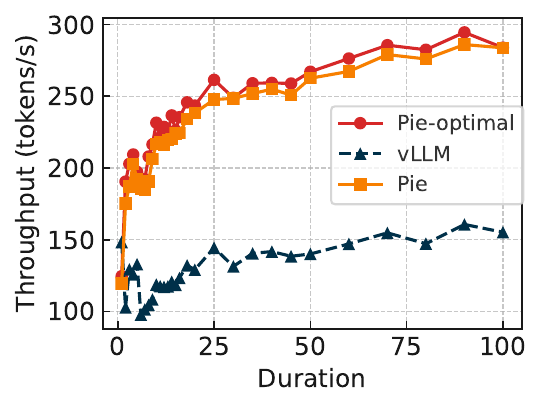}
        \caption{Adaptive Expansion Quickly Converges}
        \label{subfig:adaptive-expansion-converge}
    \end{subfigure}
    \hfill
    \begin{subfigure}[b]{0.31\textwidth}
        \includegraphics[width=\linewidth]{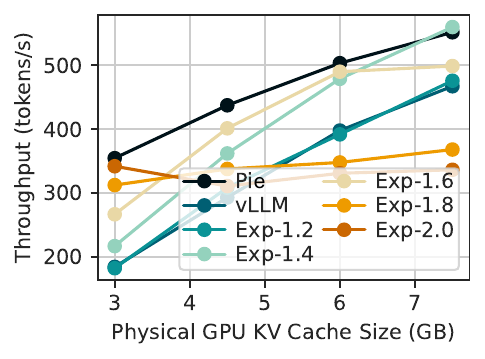}
        \caption{Adaptive Expansion Achieves Optimal Value}
        \label{subfig:adaptive-expansion-compare}
    \end{subfigure}
    \hfill
    \begin{subfigure}[b]{0.31\textwidth}
        \includegraphics[width=\linewidth]{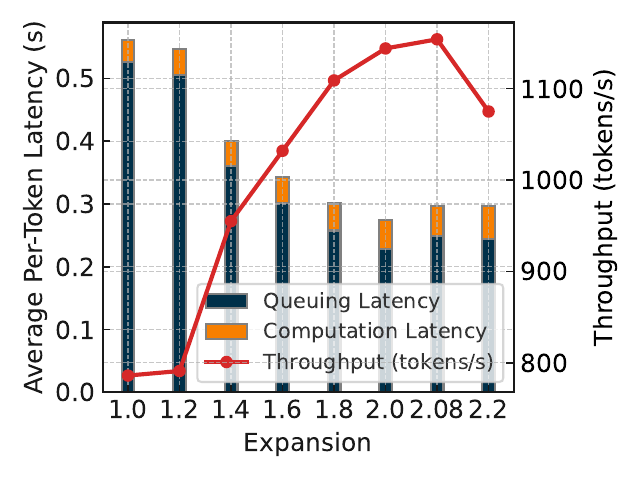}
        \caption{Higher throughput with lower latency} 
        \label{subfig:highertput-lowerlatency}
    \end{subfigure}
 
    \caption{Effectiveness of adaptive expansion. \sysname{} achieves higher throughput with lower latency.} 
    \label{fig:expansion-and-tput}
\end{figure*}

In this evaluation, we explore the performance of \sysname{} versus vLLM. 
Both systems utilize the same amount of GPU physical memory dedicated to the KV cache. By fixing the request rate, we gradually increase the physical memory size to observe the performance impact. The critical point of comparison is reached when the physical memory is sufficiently large for vLLM to handle all requests without needing to preempt and recompute. Our findings demonstrate that \sysname{} consistently outperforms vLLM by efficiently managing memory through performance-transparent swapping, thereby offering better performance even under constrained memory conditions. 

In Figure~\ref{fig:perf-over-vllm}, we vary the physical GPU KV cache size from 3GB to 12GB, where \sysname{} outperforms vLLM at smaller GPU KV cache sizes and achieve similar performance at the largest memory capacity (12GB). 
When GPU memory capacity is small, vLLM suffers from significant recomputation. During this period, capacity expansion has a crucial impact on performance. 
\sysname{} outperforms vLLM by up to 1.5$\times$ on Alpaca dataset, and $1.9\times$ on ShareGPT dataset, across different Physical GPU KV cache sizes. 
As more memory becomes available on the GPU, the improvement ratio gradually decreases until \sysname{} and vLLM perform similarly.

\subsection{Scale with Request Rate}
\label{request-rate-scale}

A larger effective cache size allows the cache to handle a higher request rate. 
We find that increasing the expansion enables \sysname{} to effectively scale the throughput when the request rate is able to saturate the KV cache size. On Alpaca dataset, \sysname{} is able to achieve $1.2-1.6\times$ throughput improvement. On ShareGPT dataset, \sysname{} is able to improve throughput by $1.6-1.9 \times$.  This is due to \sysname{}'s performance-transparent swapping which enables \sysname{} to expand its GPU KV cache size to effectively utilize on-CPU memory, enabling larger effective batch size and hence higher throughput. \sysname{}'s performance-transparent swapping achieves minimal overhead with near-linear scaling of throughput with respect to expansion, when the KV cache is saturated.  
We observe that under low request rate (e.g. $< 15.0$ requests per second for Alpaca, and $< 3.0$ requests per second for ShareGPT), expansion has less effect on the request throughput as the arrival request rate is unable to saturate the KV cache.

\subsection{Adaptive Expansion}
\label{adaptive-expansion}

In this section, we show adaptive expansion's effectiveness.
Figure~\ref{subfig:adaptive-expansion-converge} shows \sysname{} quickly adjusting its expansion value to an optimal level. \sysname-optimal uses the best expansion value from exhaustive testing, keeping CPU-GPU bandwidth fully utilized without delays. Compared to \sysname-optimal, \sysname{} incurs a less than 2\% overhead due to online monitoring and adaptation.

At the beginning of a workload, \sysname{} needs some time to adapt to the current model/workload/hardware combination. But it shortly finds the optimal value of expansion.
For most workloads, the time it takes for \sysname{} to achieve optimal expansion is within the cache warm-up time, having negligible overhead as inference services are typically long-running.

As shown in Figure~\ref{subfig:adaptive-expansion-compare}, \sysname{} achieves optimal throughput compared to swapping with other expansion values (from 1.2 to 2.0). \sysname{} is able to adaptively configure expansion across different KV cache sizes, workload, and models. We observed similar trends in other datasets and model combinations.

\subsection{Lower Latency, Higher Throughput}
\label{latency-throughput}

Figure~\ref{subfig:highertput-lowerlatency} demonstrates how \sysname{} achieves higher throughput with lower per-token average latency. Per-token latency consists of queuing and computation latency.
By using \sysname{} and increasing the KV cache memory size through the expansion value, we found that computation latency remains relatively constant, while higher expansion values reduce queuing latency. This trend is shown in Figure~\ref{subfig:highertput-lowerlatency}, illustrating improvement from expansion 1.0 to 2.0.

If the expansion value is too high, swapping performance transparency may be compromised. The latency of transferring the extended memory scales linearly with its size, eventually exceeding computation latency. This causes computation to wait for data, increasing its latency. Consequently, queuing latency also rises because, despite the larger effective capacity, each token's longer computation time delays tokens in the queue. This is illustrated by expansion 2.2 in Figure~\ref{subfig:highertput-lowerlatency}, where both compute and queuing latencies are longer compared to expansion 2.0 due to this blocking effect.

Throughput improves with larger effective memory size, though longer compute latency can compromise this improvement.
Optimal throughput is achieved with an expansion value of 2.08, which is obtained through the adaptive expansion. While cache reallocation increases queuing latency, the larger effective capacity compensates for this, resulting in higher overall throughput.

\subsection{Comparison With FlexGen}
\label{pi-vs-flexgen}

\begin{figure}[htbp]
    \centering
    \begin{subfigure}[b]{0.23\textwidth}
        \includegraphics[width=\linewidth]{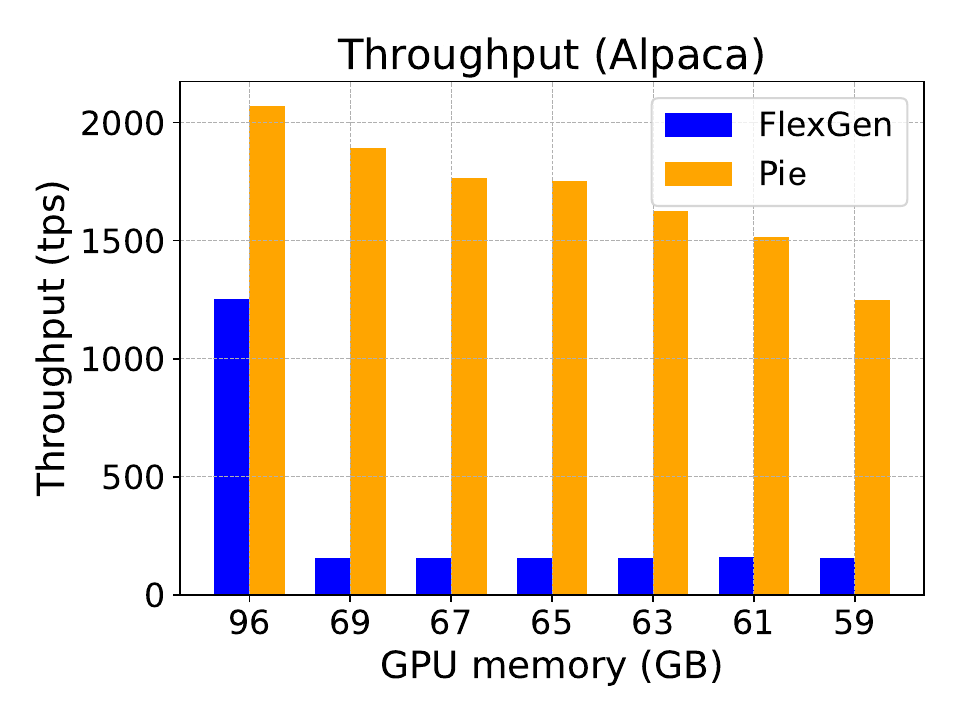}
        \caption{Throughput (Alpaca)}
        \label{throughput-flexgen-alpaca}
    \end{subfigure}
    \hfill
    \begin{subfigure}[b]{0.23\textwidth}
        \includegraphics[width=\linewidth]{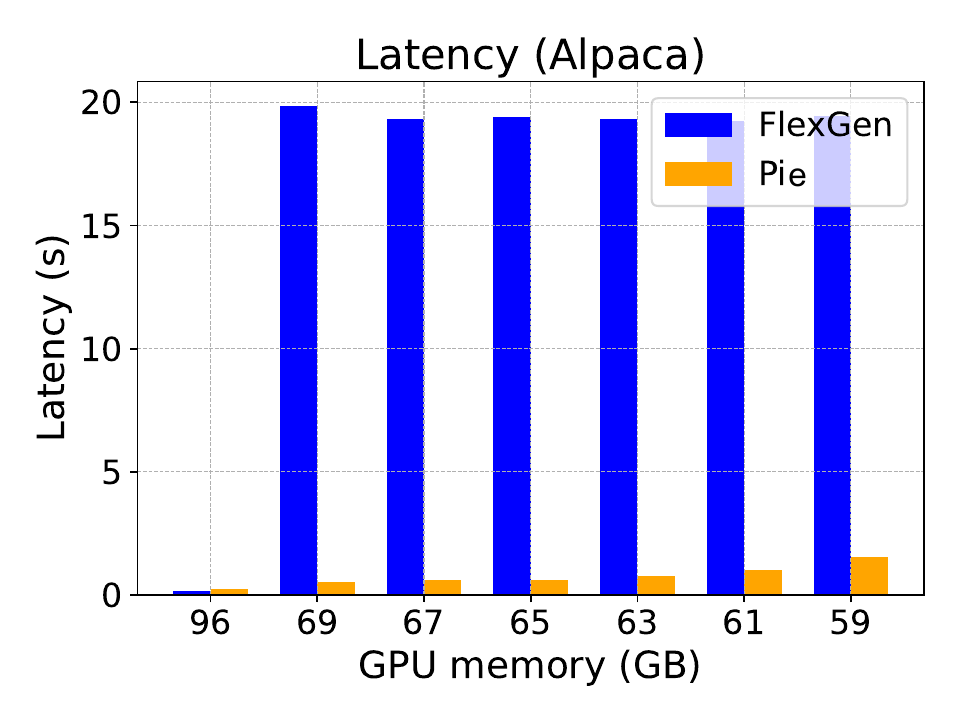}
        \caption{Latency (Alpaca)}
        \label{latency-flexgen-alpaca}
    \end{subfigure}

    \begin{subfigure}[b]{0.23\textwidth}
        \includegraphics[width=\linewidth]{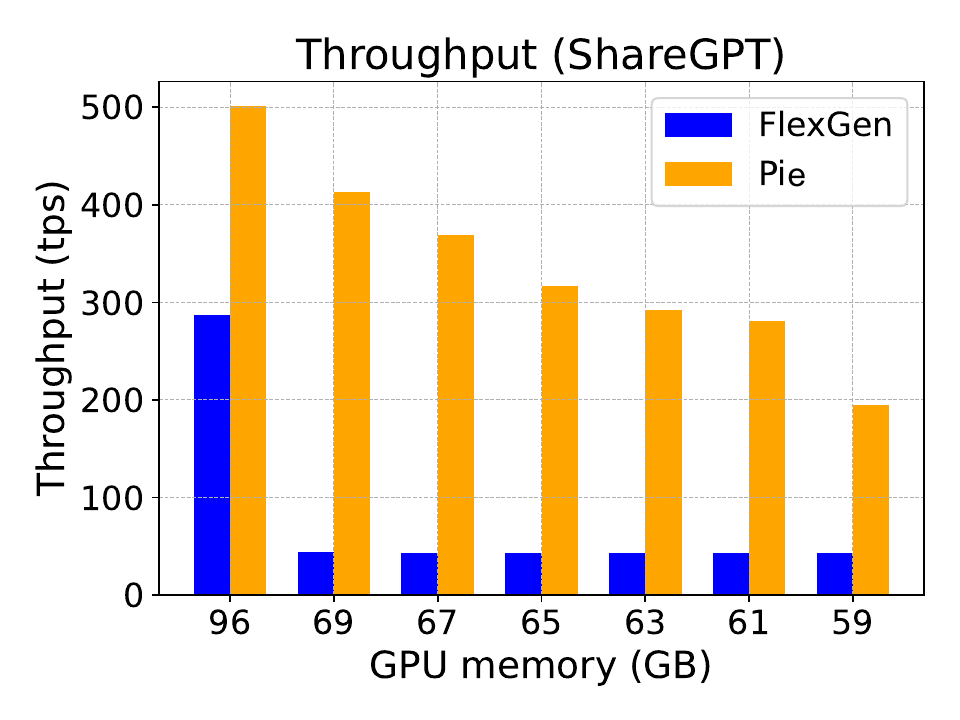}
        \caption{Throughput (ShareGPT)}
        \label{throughput-flexgen-sharegpt}
    \end{subfigure}
    \hfill
    \begin{subfigure}[b]{0.23\textwidth}
        \includegraphics[width=\linewidth]{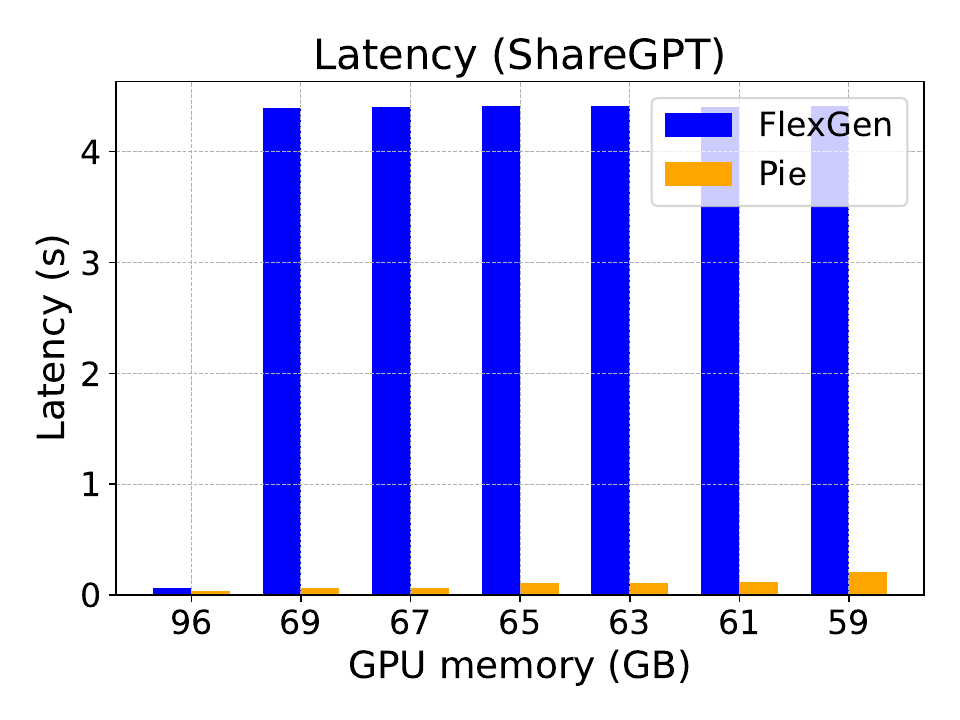}
        \caption{Latency (ShareGPT)}
        \label{latency-flexgen-sharegpt}
    \end{subfigure}

    \caption{Comparison of \sysname{} and FlexGen performance on Alpaca and ShareGPT datasets in terms of throughput and latency.}
    \label{flexgen-comparison}
\end{figure}

We evaluated the OPT-30B model using the ShareGPT and Alpaca datasets on both \sysname{} and FlexGen. \sysname{} can handle requests of varying sizes, whereas FlexGen requires all input sequences to be the same length, needing manual padding to the maximum length for each dataset.

For each configuration, we provided FlexGen's cost model with the available GPU and CPU memory sizes as input and run its generated policies on GH200. The cost model is fit with profiling data points from different weight, activation, and attention percentage setups, different number of GPU batches (e.g. 3, 8, 12, 24), different GPU batch sizes (e.g. 32, 64, 128, 256), and prompt length (e.g. 20, 162), and generation length (e.g. 32, 64, 128, 256, 512). In total, we profile 120 configurations to fit the cost model, which takes hours to finish all runs. The results are shown in Figure~\ref{flexgen-comparison}. 

When the GPU memory size is less than 96GB, FlexGen generates a policy that offloads weights to the CPU memory. This significantly reduces throughput and increases generation latency. On Alpaca, FlexGen's throughput is 7.8 $\times$ worse than \sysname{}. On the ShareGPT dataset, which has a longer prompt length and larger generation output length, \sysname{} achieves 9.4 $\times$ better throughput than FlexGen. 
It is also important to note that \sysname{} only uses the effective extended memory size of the CPU, which is less than 15GB in all setups, whereas FlexGen assumes it can freely use the entire 480GB of available CPU memory when generating its policies.
\section{Related Work}
\label{sec:related}

\paragraph{Generic Model Serving Systems}: These systems aim to improve performance, scalability, and flexibility~\cite{li2023alpaserve, crankshaw2017clipper, olston2017tensorflow, shen2019nexus, crankshaw2020inferline, zhang2023shepherd, yu2022orca}. 
FasterTransformer, LightSeq, TurboTransformers, DeepSpeed-Inference, and Hugging Face Accelerate focus on transformer optimization~\cite{nvidia_fastertransformer, wang2020lightseq, fang2021turbotransformers, aminabadi2022deepspeed, huggingface_accelerate}.
They have achieved significant improvements, but are still constrained by GPU resource limitations.

\paragraph{Solving Memory Scarcity Through Recomputation and Swapping}: 
Many works have proposed trading additional recompute time for reduced memory usage and performance improvement~\cite{chen2016training, herrmann2019optimal, korthikanti2023reducing, jain2020checkmate, dao2022flashattention, dao2023flashattention}, but most of these approaches focus primarily on training.

Swapping is another widely used technique to address memory scarcity, also with many proposals focused on training~\cite{ren2021zero, huang2020swapadvisor, rhu2016vdnn, wang2018superneurons, peng2020capuchin, yang2024protrain}.
vLLM~\cite{kwon2023efficient} swaps at the request level, causing interruptions in computation for swapped-out requests and providing no guarantee against delays in serving due to data transfers.
FlexGen~\cite{sheng2023flexgen} swaps in data only at the time of access, leading to significant delays.
Infinite-LLM~\cite{lin2024infinite} orchestrates all available GPU and CPU memories across the data center to store the KV cache. While it does not employ explicit swapping operations, applications may experience lower performance when the data is stored in CPU memory or remote memory.
InferCept~\cite{abhyankarinfercept} focuses on augmented LLMs and efficient interception to reuse previously generated context, employing pipeline swapping mechanisms. However, its naive pipeline can still block computation, with no attempts to guarantee the data will be ready when needed.

These methods overcome memory limitations, allowing for larger models and complex workloads in resource constrained environments but may compromise performance by exposing lower performance devices.

\section{Conclusion}
\label{sec:conclusion}

In conclusion, \sysname{} effectively expands GPU memory size through performance-transparent swapping and adaptive expansion. By leveraging predictable memory access patterns, modern hardware's high bandwidth, and real-time online information, \sysname{} optimizes the memory expansion ratio, ensuring efficient GPU and CPU memory use while maintaining low computation latency and high throughput across varying conditions. Our evaluation demonstrates that \sysname{} achieves high throughput, low latency, and quickly adapts to environments.

\section*{Acknowledgements}

This research is partly supported by gifts from DeepInfra, Lambda Labs, and NVIDIA, who provided NVIDIA Grace Hopper (GH200) instances for our experiments. We thank them for their support.

\bibliographystyle{plain}
\bibliography{reference}

\end{document}